# Religious Affiliation in the Twenty-First Century: A Machine Learning Perspective on the World Value Survey

Elaheh Jafarigol[1*], William Keely[1†], Tess Hortag[1†], Tom Welborn[1†], Peyman Hekmatpour[2†], Theodore B. Trafalis[3]

[1*]Data Science and Analytics Institute, University of Oklahoma, 202 W. Boyd St., Room 409, Norman, 73019, Ok, USA.
[2]Department of Sociology, University of Oklahoma, 780 Van Vleet Oval, Kaufman Hall, Room 331, Norman, 73019, OK, USA.
[3]Industrial and Systems Engineering, University of Oklahoma, 202 W. Boyd St., Room 104, Norman, 73019, OK, USA.

*Corresponding author(s). E-mail(s): elaheh.jafarigol@ou.edu;
Contributing authors: william.r.keely@ou.edu; tesshartog@ou.edu; thomaswelborn@ou.edu; peyman.hekmatpour@ou.edu; ttrafalis@ou.edu; [†]These authors contributed equally to this work.

**Abstract**

This paper is a quantitative analysis of the data collected globally by the World Value Survey. The data is used to study the trajectories of change in individuals' religious beliefs, values, and behaviors in societies. Utilizing random forest, we aim to identify the key factors of religiosity and classify respondents of the survey as religious and non-religious using country-level data. We use resampling techniques to balance the data and improve imbalanced learning performance metrics. The results of the variable importance analysis suggest that Age and Income are the most important variables in the majority of countries. The results are discussed with fundamental sociological theories regarding religion and human behavior. This study is an application of machine learning in identifying the underlying patterns in the data of 30 countries participating in the World Value Survey. The results from variable importance analysis and classification of imbalanced data provide valuable insights beneficial to theoreticians and researchers of social sciences.

## 1 Introduction

World Values Survey (WVS) is a non-commercial, non-governmental program dedicated to education and research. WVS is a publicly available dataset from a global survey since 1981. WVS aims to assess the population of the participating countries on all continents, on different criteria, and to identify what people most value in life, their impact on cultural stability, social and political issues, and their development over time [1]. The survey data is available in the form of cross-national and time series. The data consists of 7 rounds of surveys between 1981 and 2022 in different countries and among different ethnic groups. In this study, we use machine learning to explore the data and answer two questions:

1. What are the strong indicators of religious beliefs and practice of such doctrines among the population?
2. Can we utilize machine learning to classify people as religious or non-religious using historical data?

Machine learning is an umbrella term for a group of algorithms used to identify the underlying patterns present in the data in an iterative process [2]. Machine learning involves statistical analysis, regression, studying the correlations between variables and their impact on the dependent variable, feature important analysis and feature selection, methods to address data-related issues such as resampling, and algorithms like random forest for classification. A detailed discussion of the machine learning methods used for data exploration and classification is presented in sections 2 and 3, respectively. The WVS data evaluates several aspects of religiosity, such as involvement in religious services and attachment to religious beliefs. In this section, we review some of the relevant terms and theories in the sociology domain that we will use later in the discussion of our results. Secularization theory is a sociological doctrine that claims religion, in its traditional sense, is in a terminal decline as the world moves towards industrialization and economic growth. Religion has a less significant role in shaping the individual's values, beliefs, identification, and behaviors [3, 4]. The proponents of secularization theory argue that scientific advances, urbanization, and mass education decrease the influence of religion in societies. [5–8]. The rational choice theory states that individuals' decisions are based on calculations to optimize their outcomes, utility, and satisfaction. Based on this theory, personal interests and goals are the main drives of human decision-making. Rational choice theory is an influential perspective that has been applied to a wide range of social and economic phenomena. The supporters of the rational choice theory believe that religion follows the same principles of the free market and individuals decide on their level of commitment to religion while considering the costs and benefits, and that is why religious entities still exist in the 21st century [9–11]. Drawing from human's desire for stability and security, existential security theory studies the environmental, social, and economic threats to human security. Such threats drive societies towards collective actions that affect individuals and societies in positive or negative ways. Positive impacts are forming communities that lead to growth, development, and an increase in the overall well-being of society. Negative impacts are tendencies towards extreme ideologies, conflict, and sometimes violence. Focused on the patterns of change in social and political structures, this theory also argues that lack of financial security causes religion to be more present in an individual's life [12]. Social stratification is the inherent non-uniform distribution of wealth, resources, power, and social status within societies. Inequality results from unequal stratification of resources based on class, race, and gender that affects income, education, health, etc. [13]. High levels of stratification in society prevent minorities from access to resources and opportunities leading to increasing poverty among minority groups [14, 15]. Researchers such as Norris and Inglehart [16], Gorski et al.[17], and Stark et al.[18] have studied the ramifications of changes in societies and religiosity. Over the years, numerous studies have investigated the environmental factors shaping personal values and religious beliefs[19]. However, critiques such as Vasquez [20, 21] argue that more qualitative and quantitative analysis of nationally representative data is required to verify the theory on global data. Therefore, this study quantitatively analyzes WVS data using data-driven techniques to support the theories shown in Figure 1.

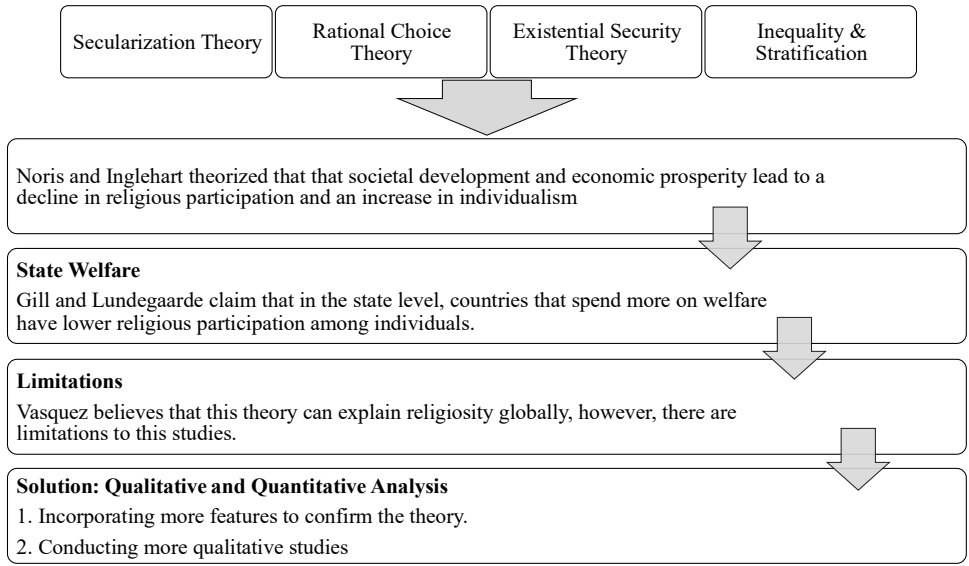

**Fig. 1**: Identifying the research question

We aim to investigate the linear and non-linear patterns in the data to identify the key indicators of religiosity in the global context and predict religiosity.

The remainder of the paper is structured as follows: section 2 presents the exploratory data analysis and a detailed discussion of the results of feature importance analysis using random forest. Subsequently, section 3 presents the results from the classification model using random forest and addresses the issue of imbalanced data. We investigate the challenges and limitations of imbalanced data and explore multiple resampling methods to improve classification performance. Finally, Section 4 concludes the study and makes suggestions for future research directions.

## 2 Analysis

### 2.1 Overview of the Data

In this study, we analyze 440055 observations collected from cross-national respondents participating in the global survey collected between 1989-2022. The respondents self identify in response to survey questions, and the subjects are selected using stratified sampling to ensure that all individuals in the population are equally likely to be selected. WVS is a valuable source for studying cultural and social developments in societies[22, 23]. In this study, religiosity is measured by the frequency of attendance at religious services and community events as part of the commitment to the practice of religious beliefs. The dependent variable, religiosity, is constructed on the basis of three items related to religion; salience, identification, and behavior, as suggested by Welzel et al. [24]. The religiosity score is calculated based on the following survey questions. The procedure for calculating the religiosity score is explained in detail in the paper published by Hekmatpour et al.[25].

1. How important is religion in your everyday life?
2. Are you a religious person?
3. Apart from weddings and funerals, how often do you attend religious services?

The response values are scaled and combined to get a continuous value between 0 and 1 as the religiosity score, while 0 means *Not religious*, and 1 is *Highly religious*. In this study, the variables are converted to 0 and 1 for binary classification. The independent variables are age, education, employment status, gender, income, marital status, life satisfaction, political view, and social class. Table 1 shows the factor levels of independent categorical variables in the data.

**Table 1**: Independent variables

| Variable | Levels |
|---|---|
| Education | Pre-primary |
| | Elementary |
| | Secondary |
| | Tertiary |
| Life Satisfaction | Very satisfied |
| | Satisfied |
| | Neutral |
| | Unsatisfied |
| | Very unsatisfied |
| Political View | Left |
| | Left-leaning |
| | Moderate |
| | Right-leaning |
| | Right |
| Social Class | Lower class |
| | Working class |
| | Middle class |
| | Higher class |

An important variable in understanding religiosity is the geographical region where the participants reside. The underlying cultural, geopolitical, and historical characteristics of the countries impact religiosity. Studying the complete list of countries present in the survey is beyond the scope of the paper. Therefore, we have limited the study to 30 countries with an adequate sample size for the classification task. The countries are selected from different regions with various cultural, political, and socio-economic structures.

In this study, we focus on the country-level data for the feature selection, resampling, and classification tasks. The models are trained on the data from each country and fine-tuned to produce the most accurate results. Since the data collected from the countries are studied separately, the results from the analysis are not impacted by the data from other countries. This is crucial in limiting the possible biases as a result of limiting the study to 30 countries rather than all the participants in the survey. It is worth mentioning that the interpretations of the results from the feature importance analysis are limited to each country and can not be extrapolated to the whole dataset. Further analysis of the remaining participating countries of the survey is required to gain a global view of religiosity.

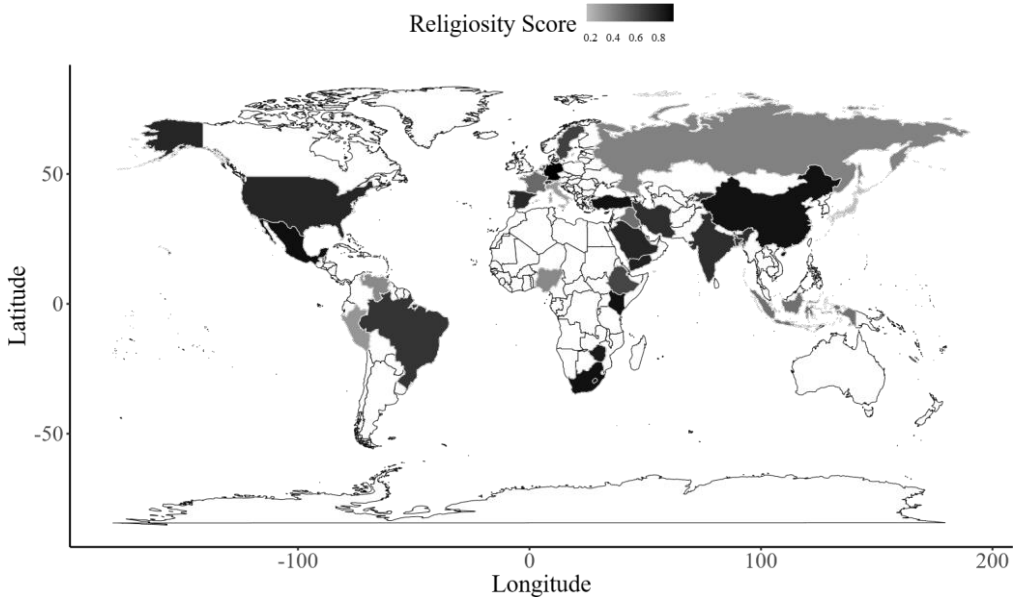

**Fig. 2**: World map of the selected countries and their average religiosity score

Figure 2 shows the world map with the selected countries and their average religiosity score which varies significantly between countries. The number of missing values in the country-level data was not significant, and we kept them labeled as *Unknown* to avoid losing valuable information. While the religiosity score varies between countries, it changes over time as well. Figure 3 shows the religiosity score between 1995 and 2022 with a 95% confidence interval shown in the gray shade. The curve shows that the most recent surge in religiosity started after 2019. Personal beliefs and values are the core contributors to religiosity. However, different global and national events can affect the rise and fall in religiosity scores. In this study, we focus on individual-level data and the factors that contribute to higher religiosity scores.

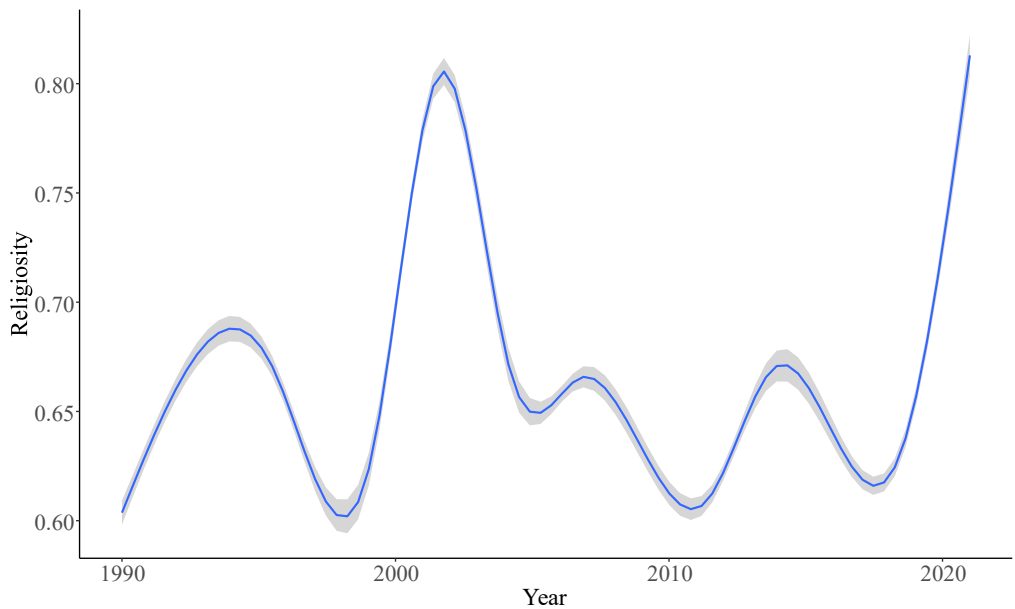

**Fig. 3**: Religiosity changes over time (1995-2022)

## 2.2 Feature Selection

This section focuses on the use of quantitative analysis to rank the features and select the ones that have a greater impact on religiosity as the target variable. We have then used the literature on religiosity in social sciences to validate the findings of our analysis and interpret the results. We used the random forest model to evaluate and rank the independent variables by their importance in predicting the religiosity score and classifying individuals based on their scores. Random forest is used for classification and regression in different domains. A random forest is a collection of hierarchical structures with a set of rules for dividing large, diverse data into smaller homogeneous groups with respect to a target variable. Random forest uses decision trees to reduce error by constructing a multitude of decision trees in parallel during training and picking random splits in the trees. It outputs a prediction that is the mean or the majority vote of the individual trees. Random forest is a powerful method with interpretable results that can handle missing values and noisy data. Therefore, it is selected as the method of choice in this study. The splitting criterion in the decision trees is the measure of impurity, which is the measure of the homogeneity of the target variable. Node impurity is the average decrease in the impurity as a result of splitting the data based on the values of the variable over all the trees in the forest. The variables are ranked from the highest increase in node impurity to the lowest, the highest being the most important. The training continues until the impurity at each node is minimized and the highest accuracy is achieved. In this study, We use the metric *Increase in Node Purity* calculated in R. To start the analysis, we have provided some statistics about the country-level data in Table 2. Figure 4 shows the distribution of religiosity in each country.

**Table 2**: Summary statistics of religiosity in the selected countries

| Country | Count | Mean | Standard Deviation | Median |
|---|---|---|---|---|
| Nigeria | 7359 | 0.93 | 0.13 | 1 |
| Ethiopia | 2610 | 0.89 | 0.17 | 0.94 |
| Kenya | 1224 | 0.88 | 0.16 | 0.94 |
| Zimbabwe | 3530 | 0.88 | 0.18 | 0.94 |
| Bangladesh | 3873 | 0.88 | 0.15 | 0.94 |
| Indonesia | 5517 | 0.88 | 0.15 | 0.94 |
| Rwanda | 2933 | 0.84 | 0.16 | 0.89 |
| Yemen | 916 | 0.81 | 0.18 | 0.78 |
| South Africa | 10864 | 0.78 | 0.24 | 0.89 |
| Iran | 5736 | 0.78 | 0.21 | 0.83 |
| India | 11090 | 0.77 | 0.21 | 0.83 |
| Brazil | 7229 | 0.75 | 0.24 | 0.83 |
| Peru | 6167 | 0.74 | 0.22 | 0.78 |
| Iraq | 6470 | 0.72 | 0.21 | 0.67 |
| Saudi Arabia | 1227 | 0.71 | 0.22 | 0.72 |
| Italy | 645 | 0.71 | 0.25 | 0.78 |
| Turkey | 10259 | 0.71 | 0.25 | 0.78 |
| Mexico | 8822 | 0.70 | 0.27 | 0.78 |
| Venezuela | 3298 | 0.67 | 0.26 | 0.72 |
| Kyrgyzstan | 3607 | 0.66 | 0.24 | 0.67 |
| United States | 9845 | 0.63 | 0.33 | 0.72 |
| Switzerland | 1951 | 0.49 | 0.32 | 0.56 |
| Spain | 4855 | 0.46 | 0.34 | 0.50 |
| Russia | 8206 | 0.46 | 0.29 | 0.56 |
| Germany | 5998 | 0.39 | 0.33 | 0.33 |

| France | 870 | 0.36 | 0.31 | 0.33 |
| Netherlands | 2214 | 0.36 | 0.34 | 0.33 |
| Japan | 5679 | 0.31 | 0.24 | 0.22 |
| Sweden | 3882 | 0.30 | 0.28 | 0.22 |
| China | 6129 | 0.16 | 0.24 | 0.11 |

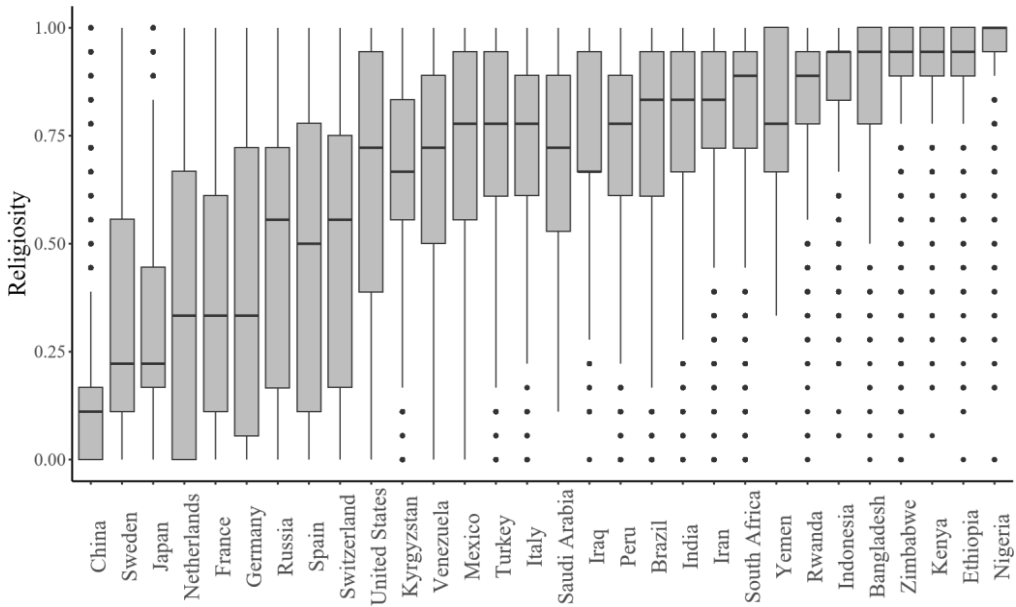

**Fig. 4**: Distribution of religiosity in the selected countries

Geographical background plays a critical role in the dispersion of religious beliefs and values. It can be observed from the figure that most boxes are roughly symmetric. The exceptions to this are Iraq, Indonesia, and Nigeria, where the data is highly skewed, such that the median falls outside the range of the box. In this case, the whisker on the longer side of the box is likely to be much longer than the whisker on the shorter side, indicating the direction of the skew. This means that more than 50% of the respondents in Iraq have a religiosity score above the median. From Table 2, we can also observe that Indonesia and Nigeria have significantly high average religiosity scores, so the data is more skewed toward higher religiosity scores among the respondents.

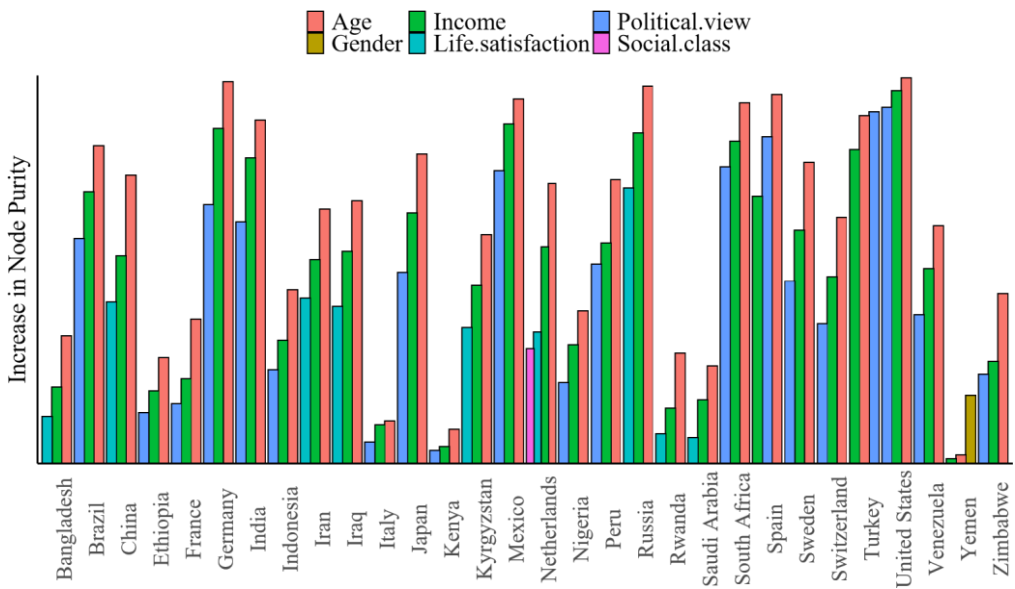

**Fig. 5**: Top three variables with the highest increase in node purity for each country

Figure 5 shows the results of variable importance analysis using random forest on country-level data. The precise magnitude of the increase in node purity is data-dependent and does not influence our interpretation or analysis of the feature importance results. While the variable importance varies in different countries, *Age* and *Income* are the top two most important variables in 93.3% of the countries. The next most important variable is *Political status* in 60% of the countries and *Life satisfaction* in 26% of the countries. In the next subsections, we further examine the four variables common among most countries that have the greatest increase in node purity.

### 2.2.1 Age

The fitted curve in Figure 6 shows the relationship between religiosity and age in the data, and Figure 7 presents the distribution of religiosity by country.

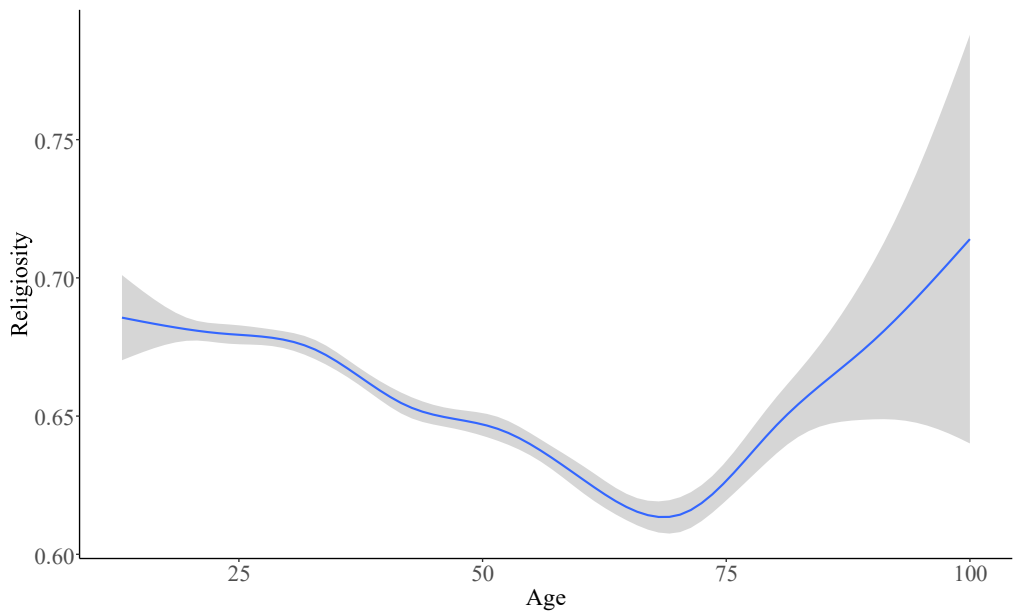

**Fig. 6**: The trend in religiosity by age: focusing on the collective data, the religiosity score is higher among elderly

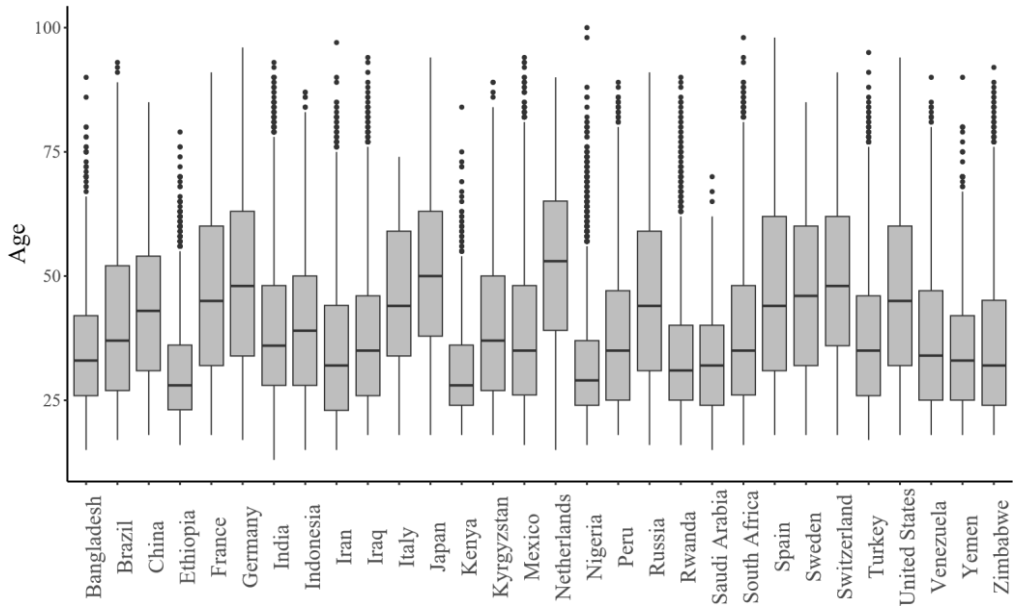

**Fig. 7**: The box plots show the distribution of participants' age in the selected countries.

It is observed that the religiosity score declines until the middle ages but increases as individuals grow older. The wide confidence interval of the fitted curve in Figure 6 indicates the variance between the religiosity scores of the elderly in different countries. Table 3 presents the results from statistical analysis and fitting a linear regression model on the country-level data.

**Table 3**: Correlation analysis and linear regression results

| Country | Correlation Coef. | Adjusted $R^2$ | P-value | F-statistic |
|---|---|---|---|---|
| China | 0.061 | 0.004 | 0.00 | 22.81 |
| India | 0.076 | 0.006 | 0.00 | 63.71 |
| Japan | 0.218 | 0.047 | $< 2.2e-16$ | 283.1 |
| Mexico | 0.168 | 0.028 | $< 2.2e-16$ | 255.2 |
| Nigeria | 0.022 | 0.000 | 0.054 | 3.71 |
| Russia | 0.127 | 0.016 | $< 2.2e-16$ | 133.40 |
| Spain | 0.343 | 0.117 | $< 2.2e-16$ | 645.70 |
| Turkey | 0.146 | 0.021 | $< 2.2e-16$ | 224.50 |
| United States | 0.154 | 0.024 | $< 2.2e-16$ | 240.60 |
| Brazil | 0.143 | 0.20 | $< 2.2e-16$ | 150.70 |
| Bangladesh | 0.116 | 0.013 | 0.00 | 52.74 |
| Peru | 0.190 | 0.036 | $< 2.2e-16$ | 229.70 |
| South Africa | 0.102 | 0.010 | $< 2.2e-16$ | 114.30 |
| South Arabia | 0.086 | 0.007 | 0.003 | 9.13 |
| Sweden | 0.202 | 0.041 | $< 2.2e-16$ | 165.40 |
| Switzerland | 0.261 | 0.068 | $< 2.2e-16$ | 142.40 |
| Venezuela | 0.144 | 0.020 | $< 2.2e-16$ | 69.36 |
| Germany | 0.193 | 0.037 | $< 2.2e-16$ | 233.00 |
| Iran | 0.153 | 0.023 | $< 2.2e-16$ | 137.10 |
| Indonesia | 0.167 | 0.028 | $< 2.2e-16$ | 157.20 |
| Zimbabwe | 0.001 | 0.00 | 0.947 | 0.00 |
| Kyrgyzstan | -0.005 | 0.00 | 0.771 | 0.08 |
| Iraq | 0.109 | 0.021 | $< 2.2e-16$ | 77.56 |

| Italy | 0.205 | 0.040 | 0.00 | 28.17 |
| France | 0.196 | 0.37 | 0.00 | 34.51 |
| Netherlands | 0.17 | 0.028 | 0.00 | 65.52 |
| Rwanda | 0.025 | 0.00 | 0.178 | 1.82 |
| Ethiopia | 0.106 | 0.011 | 0.00 | 29.55 |
| Yemen | 0.112 | 0.011 | 0.001 | 11.60 |
| Kenya | 0.070 | 0.004 | 0.014 | 6.04 |

The correlation coefficient ranges from -1 to 1 and measures the strength and direction of the correlation between age and religiosity. Positive values indicate a positive relationship, negative values indicate a negative relationship, and 0 is an indication of a weak or no correlation. Except for Kyrgyzstan, which has a very small negative coefficient, there is a positive relationship between religiosity and age. However, the strength varies, where Zimbabwe and Nigeria have the weakest correlations, and Spain and Switzerland have the strongest correlations. The reported P-values also indicate that there is a statistically significant relationship between the variables in most countries, with Zimbabwe, Kyrgyzstan, Rwanda, and Kenya being the exception.

Figure 8 presents the correlation between age and religiosity of individuals during their adulthood in 6 countries of China, Brazil, Spain, Ethiopia, Venezuela, and South Arabia.

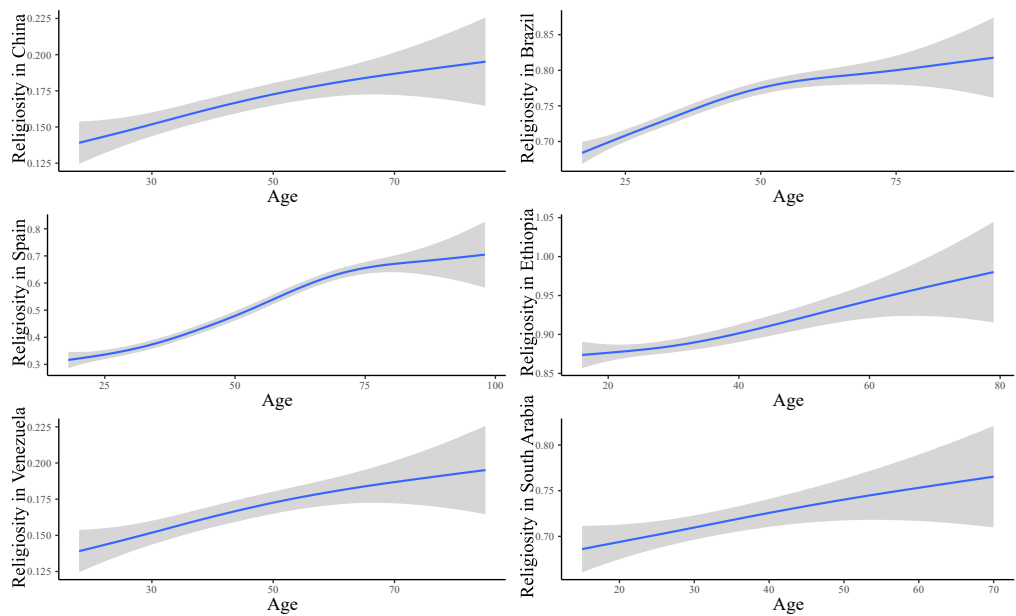

**Fig. 8**: The fitted curves show the increase in religiosity with age in participants from a sample of 6 countries present in the study. More details on the correlation between religiosity and age are provided in Table 3

Sociologists explain this behavior by existential security theory. Generally, people tend to experience more insecurity and unpredictability in their lives as they grow older. The challenges and insecurities they face as a result of aging drive people to rely on religion as a source of comfort and hope [26]. Also, psychological studies support the idea that participating in religious services creates a sense of community which improves the mental and physical health of the elderly [27, 28]. Although the high value of the F-statistic suggests that the model is a good fit for some countries, the small value of the $R^2$ shows the inability of the linear model to capture the variations of the target variable. Therefore more variables need to be included in the model to be able to fit the data and explain the variations in the target variable. In conclusion, the results from our analysis are consistent with the idea that a decrease in existential security as a result of aging impacts life choices and perspective. While these insecurities are caused by the decline in physical, cognitive, and mental abilities, health problems, financial instability, and loss of friends and loved ones, it results in an increase in religiosity with age.

### 2.2.2 Income

Figure 9 presents the negative correlation between religiosity and income. Religiosity is often higher among people with lower incomes and declines as income levels grow. While this negative correlation is true for many countries, there are also exceptions to this. Therefore, we will have a closer look at the country-level data.

In countries where the economy is unstable, lower income causes more challenges, and existential insecurity, which results in a higher tendency for religiosity. Numerous studies support this theory. The 2004 report published by the Pew Center shows that the rate of participation in religious services is higher in low-income countries, and religious beliefs are stronger[29]. Different factors contribute to the gap in income levels within societies. There is a higher level of inequality and stratification in societies with a greater gap between income levels. Economic inequality and social marginalization caused by a lack of opportunities for social mobility contribute to higher levels of stratification in societies. Consequently, low-income individuals are more inclined to participate in religious services as a source of comfort against their struggles. This theory explains the decline of religiosity in industrialized and economically stable countries versus underdeveloped countries where there is no welfare provision. Figures 10 and 11 visualize the correlations between religiosity and income in developing and underdeveloped countries and developed countries, respectively.

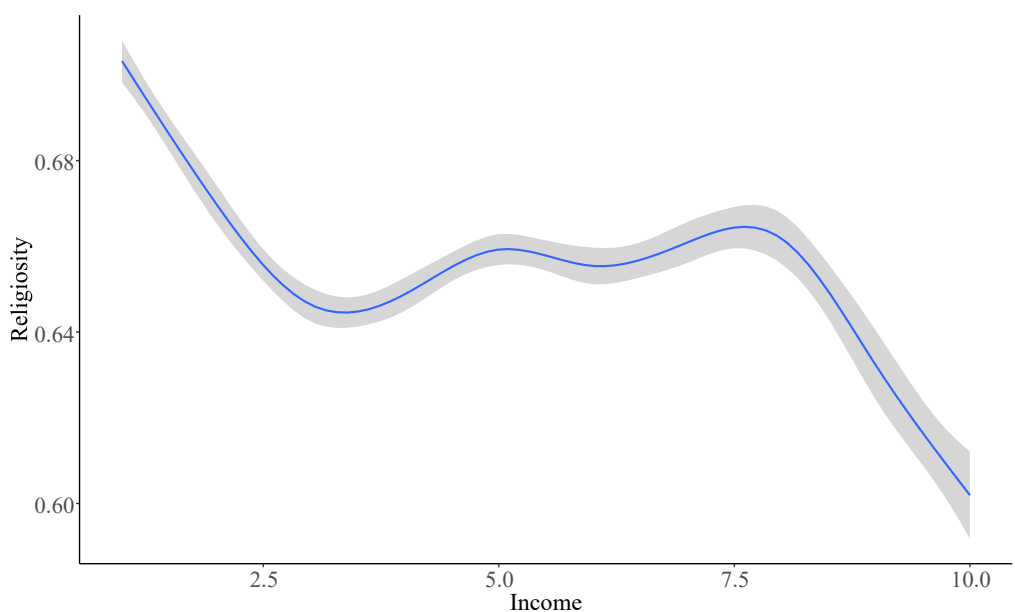

**Fig. 9**: Generally, religiosity and income are negatively correlated. A closer look at the country-level data is required to understand the variations specific to each country

Figure 10 shows the negative relationship between income and religiosity in 6 underdeveloped and developing countries of Russia, Turkey, Brazil, Bangladesh, Peru, and Iran. The decline in religiosity as a result of an increase in income levels in developing and underdeveloped countries is explained by the existential security theory caused by stratification and inequality in societies.

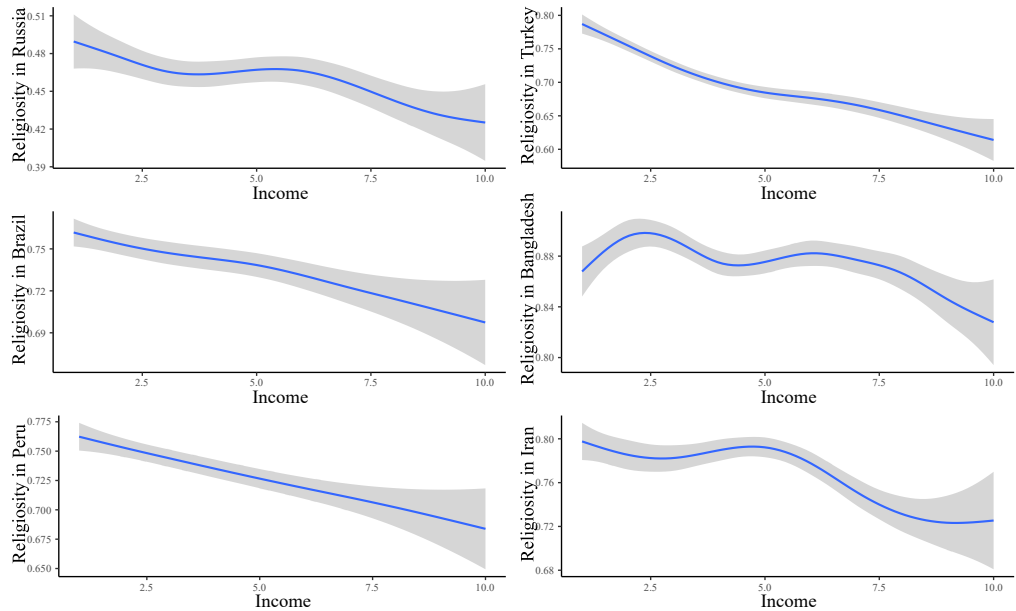

**Fig. 10**: This figure shows a sample of countries present in this study that lacks economic stability. In developing and under-developed countries, religiosity is negatively correlated with the growth in income levels

The secularization theory is another facet of the negative correlation between religiosity and income. While individuals are less likely to experience the insecurities caused by famine, disease, war, and environmental disasters, in developed counties, industrialization, and scientific growth replace religious beliefs. Therefore the overall religiosity levels decline [30, 31]. Figure 11 presents the negative correlation between religiosity and income in developed countries, and studies in sociology support this idea [32–34].

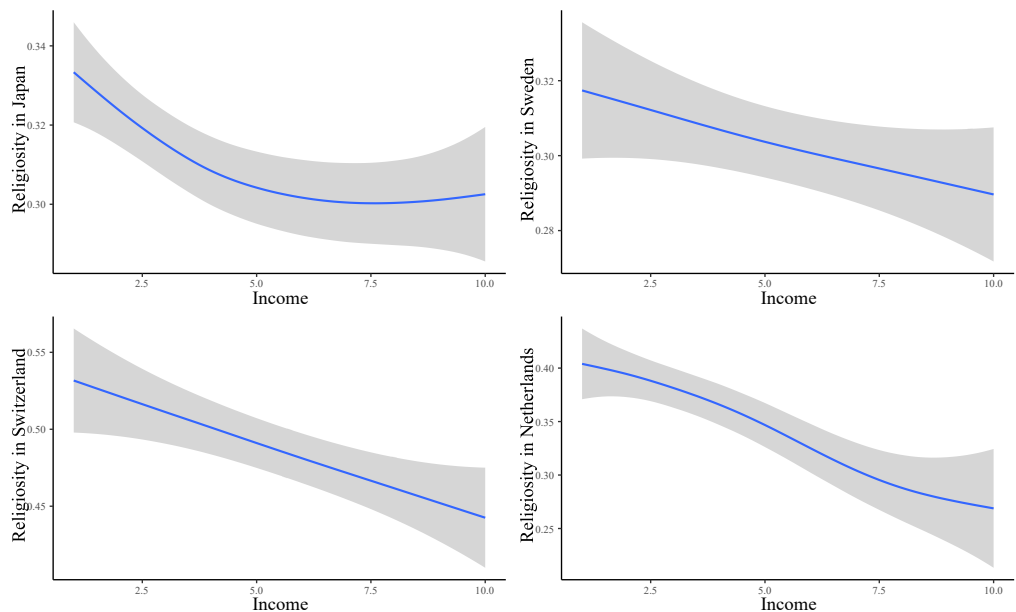

**Fig. 11**: This figure shows a sample of developed countries present in this study. In developed countries, religiosity is negatively correlated with the growth in income levels

Unlike most developed countries, there is a significant gap between various groups in the United States. Figure 12 shows the correlation between religiosity and income level in the United States. It can be observed from Figure 12 that the fitted curve has a negative slope at first and a positive slope as income levels grow. What differentiates the United States from other countries is the increase in religiosity scores after an initial drop in religiosity scores.

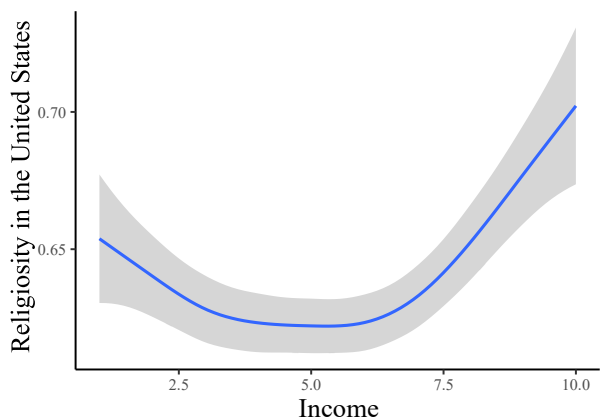

**Fig. 12**: Unlike many developed countries, religiosity and income are not always negatively correlated in the United States. In higher income levels, religiosity, and income are positively correlated

In the United States, low income is partially the result of social inequality and stratification [35]. Minority groups who do not have access to resources and opportunities endure more financial challenges and insecurities. A study published by the American Sociological Review [36] investigates the relationship between religiosity and income in marginalized and minority groups. The study shows that the inequality and stratification caused by systematic racism create an environment where individuals have limited access to resources, therefore, turn to religion for support in their communities. The study is conducted based on the data from the General Social Survey [37, 38], which also collects information about religious beliefs and practices in the US. The study concludes that participants with lower income levels have a stronger belief in God and actively participate in religious services[39]. Religiosity is a complex topic in the United States, and we will further explore the relationship between religiosity and political structure in the United States in the next subsection.

Despite the general trends, the relationship between religiosity and income is complex and affected by the cultural background, the population's beliefs, and values determined in their geographical context. Countries where religion is dominant, do not follow similar trends in religiosity as countries with secular governments. Yemen is an exception to this, where income and religiosity don't follow the same trends as other countries. In Yemen, the religiosity score doesn't consistently decrease as income levels grow, but it varies among populations with different income levels. First, the participants are categorized into low-income and high-income groups. Figure 13 shows the comparison of the distribution of religiosity between low-income and high-income populations in Yemen.

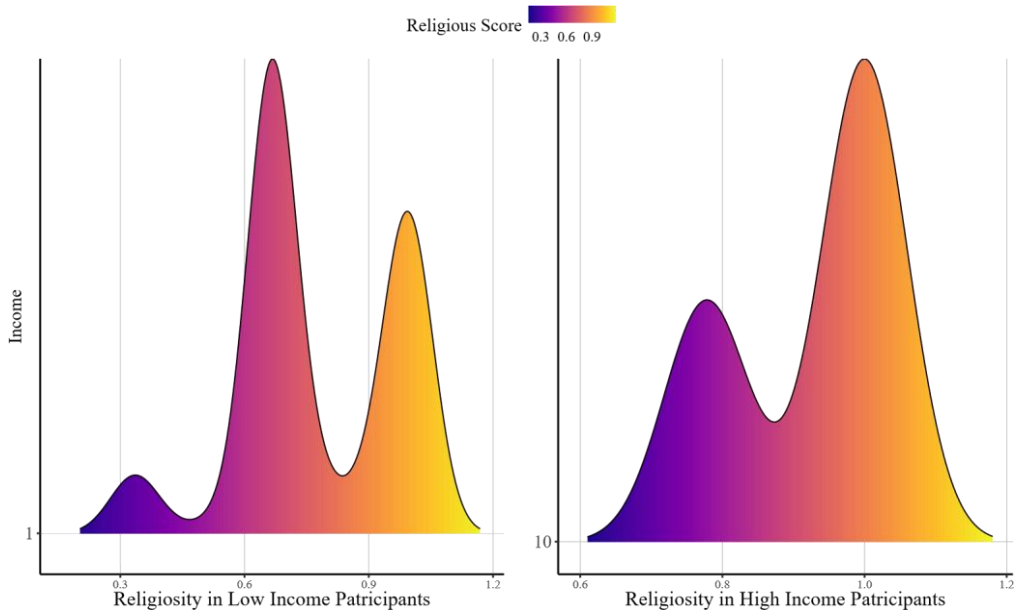

**Fig. 13**: The majority of individuals with high-income levels have higher religiosity scores. Also, the population of high-income, highly religious individuals is significantly larger than the population of individuals with high religiosity scores among low-income populations

Yemen is an Islamic country where religion is deeply intertwined with power and wealth. Islam is the dominant religion in Yemen which has a major influence on culture and social norms. Yemen's wealthy and powerful sector allocates funds and resources to support political and religious groups that align with their interests. This reflects in the spike in religiosity scores among individuals with higher income levels.

Figure 14 shows Iran, another country with a large population of religious individuals with higher income levels. What differentiates Iran from other countries is the large population of religious individuals with higher income levels. Iran is a country with an Islamic government and political system that deeply influences the social system and beliefs. In Iran, religiosity is a marker of power and the wealthy elite. Therefore, being involved in religious practices is encouraged beyond personal beliefs and values, even though it does not fully represent the population. This religious culture among the wealthy and powerful in Yemen and Iran is in line with the rational choice theory, where personal goals and satisfaction play an important role in the decision to participate in religious gatherings, even if it is in opposition to the general population. Unlike elites and people who benefit financially and socially from religious groups being in power, studies show that religiosity declines when enforced by the government, and gradually the general population becomes less religious[40, 41]. Overall, the relationship between religion and income is very complex, and it is deeply influenced by the social and political structures in society. The cultural background, the level of industrialization, equality, and equity are also essential contributors to religiosity.

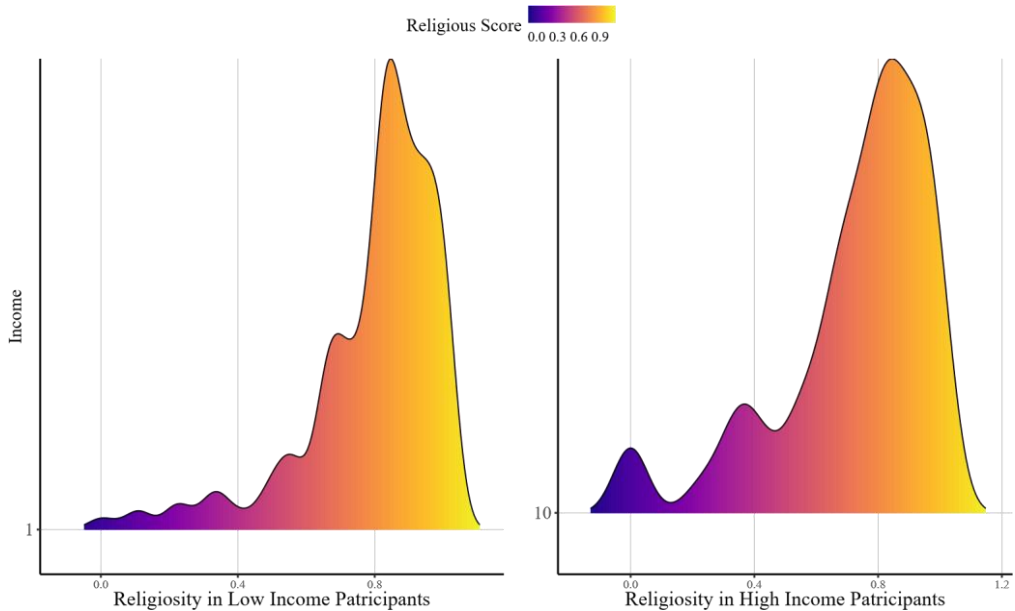

**Fig. 14**: Comparison of the distribution of religiosity between high-income and low-income populations in Iran. Iran has a substantial religious population who are present among both high-income and low-income respondents

### 2.2.3 Political Views

Religion and political views, as two components that shape individuals' understanding of the world and their place in it, are closely related. This complex relationship is deeply affected by cultural and historical backgrounds. It also varies among different political ideologies and religions. In this section, We examine religiosity in participants from developed countries with different levels of separation of religion and state. As shown in Table 2, Germany, Sweden, Netherlands, and Japan have low average religiosity scores. Figure 15 is a comparison of the distribution of religiosity at different political views in developed countries with low religiosity scores. Referring to the secularization theory, scientific advances and growth in technology have reduced the influence of religion on societies. This has caused a shift in political views and a digression from conservatism. Secularization theory suggests that people with liberal or progressive political views, such as supporters of individual rights, social equity, and environmental issues, mostly identify as less religious or non-religious. Less religious individuals tend to view the world through a secular lens. They prioritize rational and evidence-based thinking over traditional religious beliefs and values.

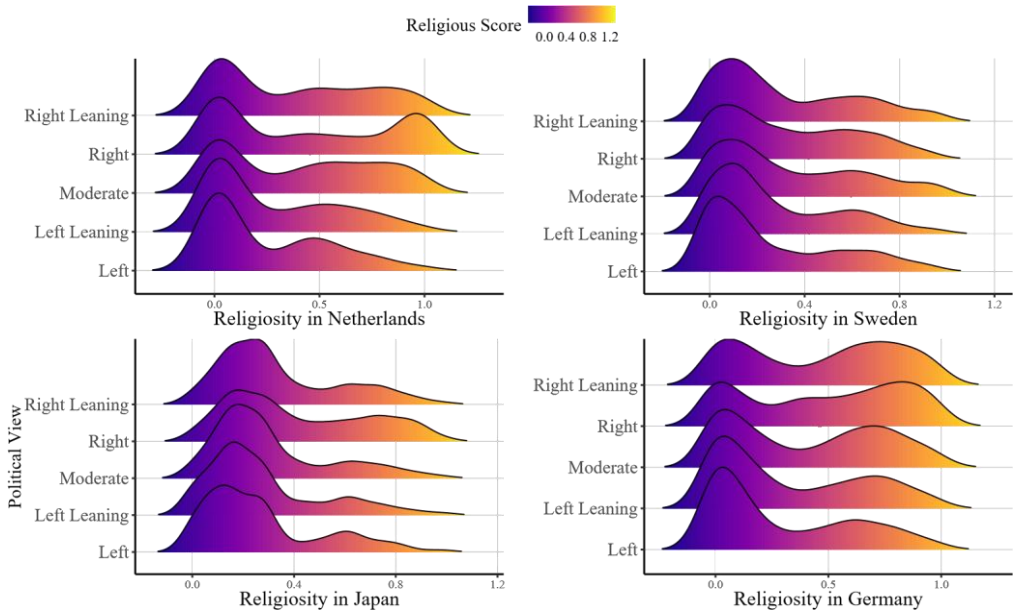

**Fig. 15**: Distribution of religiosity by political views in developed countries

In such countries, there is a greater separation of religion and state. As it is shown in Table 4, the positive and statistically significant coefficients suggest that religiosity increases as political views become more conservative. However, the correlation between religiosity and political views is weak. The small value of adjusted $R^2$ suggests the non-linear relationship between political views and religiosity in these countries.

**Table 4**: Correlation analysis and linear regression results

| Country | Correlation Coef. | Adjusted $R^2$ | P-value | F-statistic |
|---|---|---|---|---|
| Netherlands | 0.129 | 0.016 | 5.86e-09 | 34.18 |
| Sweden | 0.058 | 0.003 | 0.00 | 12.82 |
| Japan | 0.108 | 0.011 | $5.267e-13$ | 52.42 |
| Germany | 0.156 | 0.024 | $< 2.2e-16$ | 140 |

The critics of the secularization theory argue that there are various counterexamples, and this theory doesn't explain the resurgence of religious movements in some regions. Therefore alternative methods are required [4, 42]. An exception to this theory is the United States, where religiosity is a nuanced topic [43, 44]. The United States is a developed country where religious beliefs remain strong among some groups despite advances in science and technology. The comparison of the distribution of religiosity between political views in Figure 16 confirms that the majority of participants with right and right-leaning views have higher religiosity scores versus the majority of the participants with left and left-leaning views having lower religiosity scores. This idea is supported by studies in sociology showing the positive correlation between religion and American political culture [45]. Historically, the republican party in the United States has more right and right-leaning views, and the majority of the democratic party has left and left-leaning views. While conservative Christians in the United States have been more affiliated with the Republican party, liberal conservatives have been more associated with the Democratic party, which is consistent with the findings of our analysis and the rational choice theory, where religion as an entity contributes to individuals living in different ways, and vice versa [46]. In conclusion, with the rapid growth in science and technology, more countries are moving towards industrialization, so the relationship between secularization and political views is an area of ongoing research and debate as scholars seek to understand the complex interplay between religion, politics, and social change in modern societies.

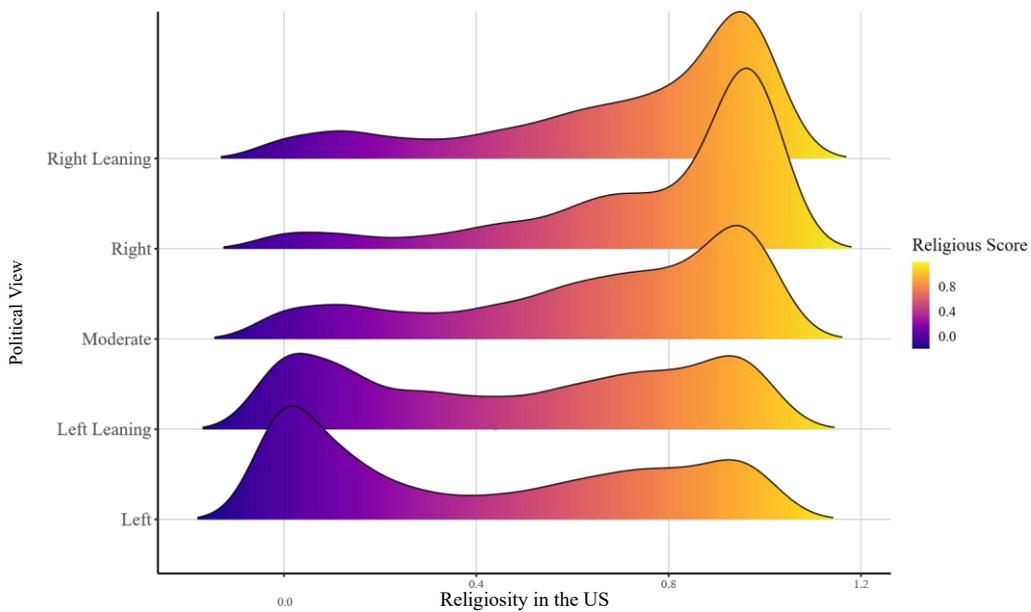

**Fig. 16**: Comparison of the distribution of political views in the US

### 2.2.4 Life Satisfaction

The last important variable based on the feature importance analysis provided in Figure 5 is life satisfaction. Figure 17 presents the distribution of religiosity in different levels of life satisfaction in countries where life satisfaction is the third most important variable. These countries are Bangladesh, China, Iran, Iraq, Kyrgyzstan, Russia, Rwanda, and Saudi Arabia, which are mainly developing and under-developed countries. Figure 17 shows that in such countries, religiosity has a positive correlation with life satisfaction, happiness, and well-being.

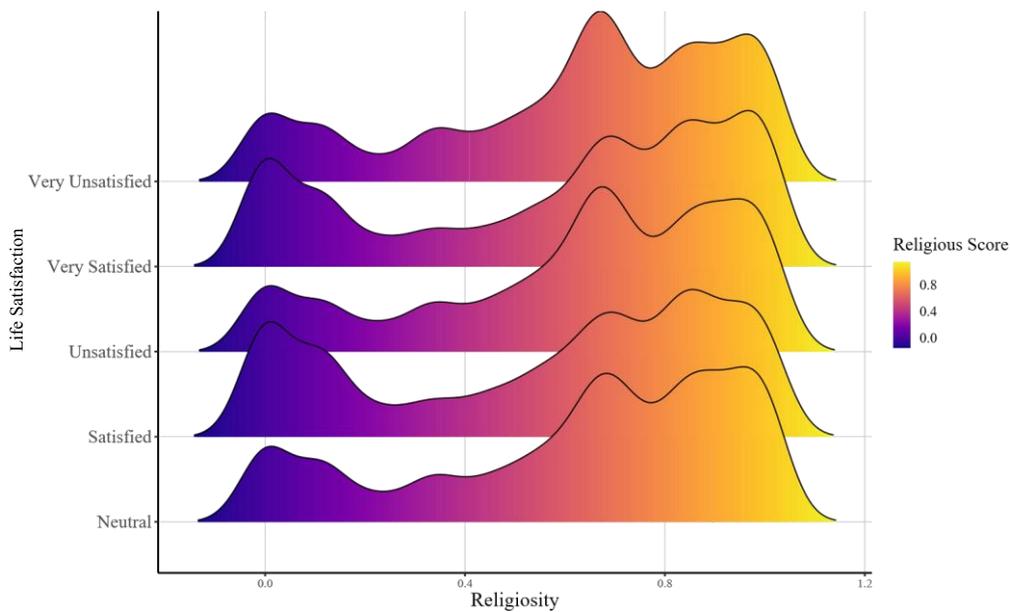

**Fig. 17**: Countries where life satisfaction is ranked as the third most important variable

In sociology, religiosity is believed to follow the principles of the free-market economic model. In this framework, societies and individuals evaluate the costs and benefits of religious commodities offered to them and choose the tradition that contributes most to their happiness and life satisfaction [42, 47]. There is evidence that suggests religiosity may be associated with higher levels of life satisfaction. Levin et al. [48] study the impact of religion on happiness and life satisfaction among Jews. Many researchers, such as Abdel-Khalek [49], have also studied the association of religious belief with happiness, physical health, and mental health. Nonetheless, to better understand the relationship between religiosity and life satisfaction, we need to consider the influence of factors such as the type of religion or spirituality in question, the cultural context, and the individual's personal beliefs and experiences.

## 3 Classification

In this section, we examine the relationships between various predictor variables and religiosity to train a machine-learning model to classify individuals as religious or nonreligious using historical data. In recent years, machine learning techniques have been increasingly used in different domains. One area of interest is in social science research to predict individuals' behavior by analyzing large datasets. By training machine learning models on historical data such as survey responses, we explore the patterns within the data and classify respondents based on their religious affiliations. We studied the key predictors and identified the factors that contribute to religiosity. In previous works, natural language processing models have been used to detect religious affiliation and religious extremism on social media [50–52], and Samantaray et al.[53] explored tree models as a feature selection technique to predict religion based on country flags. To further understand contributors to religiosity based on individuals' beliefs and practices, we utilize a random forest model to classify individuals using their responses in the WVS dataset. We explore the feasibility and accuracy of random forest for this classification task. This provides a nuanced application of key factors and their correlation with religious affiliations for categorizing religiosity levels among demographic groups and geographic locations. Religiosity is defined with a focus on the frequency of attendance in religious services and community events as an indicator of religious commitment. We start by selecting 30 countries and preprocessing the data by selecting relevant variables and encoding categorical variables. We split the dataset into training and testing and used 10-fold cross-validation to evaluate the results. The initial analysis of the country-level data showed that the data from most countries are imbalanced, meaning that the data is not equally distributed between the two classes. Class imbalance is a common issue that negatively impacts the classification performance of the model on the minority class because the model is overwhelmed by the majority class, and the results become biased. The imbalanced data combined with the small sample size causes a significant decline in model performance. In the context of classifying religiosity, it is important to have accurate predictions, especially if the cost of misclassification is high, for example, if it leads to negative consequences, lack of trust, or discrimination. We also need to monitor the overall performance of the model in predicting religiosity. To address this issue, we use metrics that incorporate performance in each class to evaluate the results. The imbalanced learning-specific evaluation metrics used in this study are accuracy, precision, recall, and geometric mean (G-mean), which are defined in Table 5.

**Table 5**: Threshold metrics in supervised learning of imbalanced data

| Metrics | Definition |
| --- | --- |
| Accuracy | The ratio of the correctly classified instances over the total number of classified instances |
| Precision | The proportion of instances that were labeled correctly among those with the positive label in the test data |
| Recall | The portion of positive instances in the test data that were labeled correctly |
| G-mean | The measure to maximize the accuracy of the model over each class by considering both classes for evaluation |

Table 6 presents the country-level data, with the proportion of minority to the majority class and the imbalanced learning metrics results on imbalanced data. The comparison of accuracy with the G-mean shows that while the accuracy is high in many countries, the models tend to perform significantly worse in the minority class. Hence the G-mean is low.

**Table 6**: Imbalanced data proportions and the classification results

| Country | Religious(%) | Non-religious(%) | Accuracy | G-mean |
| --- | --- | --- | --- | --- |
| China | 13.44 | 86.56 | 0.88 | 0.49 |
| India | 88.38 | 11.62 | 0.88 | 0.18 |
| Japan | 23.42 | 76.58 | 0.74 | 0.23 |
| Mexico | 79.69 | 20.31 | 0.80 | 0.21 |
| Nigeria | 98.60 | 1.40 | 0.99 | 0.0 |
| Russia | 56.12 | 43.88 | 0.62 | 0.60 |
| Spain | 51.51 | 48.49 | 0.71 | 0.70 |
| Turkey | 83.05 | 16.95 | 0.84 | 0.46 |
| United States | 70.33 | 29.67 | 0.73 | 0.55 |
| Brazil | 85.05 | 14.03 | 0.86 | 0.14 |
| Bangladesh | 97.99 | 2.01 | 0.98 | 0.24 |
| Peru | 86.83 | 13.17 | 0.87 | 0.10 |
| South Africa | 86.20 | 13.80 | 0.86 | 0.13 |
| South Arabia | 82.07 | 17.93 | 0.80 | 0.13 |
| Sweden | 28.57 | 71.43 | 0.285 | 0.279 |
| Switzerland | 55.20 | 44.80 | 0.57 | 0.56 |
| Venezuela | 77.62 | 22.38 | 0.78 | 0.39 |
| Germany | 44.08 | 55.92 | 0.64 | 0.60 |

| Country | | | | |
|---|---|---|---|---|
| Iran | 90.36 | 9.64 | 0.91 | 0.12 |
| Indonesia | 97.50 | 2.50 | 0.97 | 0.0 |
| Zimbabwe | 94.65 | 5.35 | 0.96 | 0.0 |
| Kyrgyzstan | 80.93 | 19.07 | 0.81 | 0.17 |
| Iraq | 85.64 | 14.36 | 0.86 | 0.19 |
| Italy | 84.34 | 15.66 | 0.86 | 0.30 |
| France | 38.62 | 61.38 | 0.96 | 0.0 |
| Netherlands | 37.94 | 62.06 | 0.69 | 0.57 |
| Rwanda | 96.73 | 3.27 | 0.66 | 0.54 |
| Ethiopia | 57.16 | 42.84 | 0.96 | 0.0 |
| Yemen | 96.59 | 3.38 | 0.97 | 0.65 |
| Kenya | 95.59 | 4.41 | 0.96 | 0.0 |

When dealing with classification, it is necessary to address the challenges of imbalanced data and small sample size. Over-sampling methods are a group of resampling techniques used for imbalanced learning. The choice of resampling method depends on the data. Therefore, three methods were used to achieve the best performance on each data set.

1. A combination of random over-sampling and random under-sampling is a method of handling imbalanced data which involves generating new samples of the minority class by selecting the k-nearest neighbors of the minority sample and creating new samples on the line joining the minority sample and one of the neighbors. Next, apply an under-sampling technique to remove the majority samples near the minority sample and reduce the overlap and noise between the two classes. The combination of over-sampling and under-sampling balances the data and improves the classification performance [54, 55].
2. Synthetic Minority Over-sampling Technique (SMOTE) is a widely used resampling technique that generates synthetic samples along the line segments that connect two samples of the minority class. SMOTE is an effective technique because it generates new instances rather than duplicating the existing ones [56, 57].
3. Adaptive Synthetic Sampling (ADA-SYN) is an extension of SMOTE which assigns a random value to the generated samples to have more variance in the synthetic data [58].

Table 7 presents the results of training the random forest on balanced data. It can be observed that the G-mean has improved substantially. The precision and recall indicate that the model is performing equally well in both classes, and high accuracy shows the overall effectiveness of the classification model.

**Table 7**: Classification results of balanced data

| Country | Accuracy | Recall | Precision | G-mean |
|---|---|---|---|---|
| China | 0.88 | 0.86 | 0.90 | 0.88 |
| India | 0.88 | 0.88 | 0.88 | 0.88 |
| Japan | 0.88 | 0.88 | 0.88 | 0.88 |
| Mexico | 0.88 | 0.88 | 0.88 | 0.88 |
| Nigeria | 0.98 | 0.97 | 0.99 | 0.98 |
| Russia | 0.82 | 0.81 | 0.81 | 0.82 |
| Spain | 0.84 | 0.81 | 0.87 | 0.81 |
| Turkey | 0.88 | 0.86 | 0.88 | 0.87 |
| United States | 0.83 | 0.85 | 0.81 | 0.84 |

| Country | | | | |
|---|---|---|---|---|
| Brazil | 0.87 | 0.88 | 0.87 | 0.87 |
| Bangladesh | 0.98 | 0.97 | 0.99 | 0.98 |
| Peru | 0.87 | 0.85 | 0.88 | 0.87 |
| South Africa | 0.86 | 0.84 | 0.89 | 0.86 |
| South Arabia | 0.86 | 0.86 | 0.87 | 0.86 |
| Sweden | 0.87 | 0.85 | 0.89 | 0.87 |
| Switzerland | 0.86 | 0.89 | 0.82 | 0.85 |
| Venezuela | 0.88 | 0.88 | 0.87 | 0.88 |
| Germany | 0.85 | 0.83 | 0.87 | 0.85 |
| Iran | 0.88 | 0.87 | 0.89 | 0.88 |
| Indonesia | 0.98 | 0.97 | 0.98 | 0.98 |
| Zimbabwe | 0.97 | 0.94 | 0.99 | 0.97 |
| Kyrgyzstan | 0.86 | 0.86 | 0.86 | 0.86 |
| Iraq | 0.90 | 0.94 | 0.88 | 0.91 |
| Italy | 0.91 | 0.88 | 0.93 | 0.91 |
| France | 0.88 | 0.83 | 0.93 | 0.88 |
| Netherlands | 0.86 | 0.82 | 0.89 | 0.86 |
| Rwanda | 0.97 | 0.97 | 0.97 | 0.97 |
| Ethiopia | 0.97 | 0.96 | 0.99 | 0.97 |
| Yemen | 0.97 | 0.95 | 1.00 | 0.98 |
| Kenya | 0.97 | 0.97 | 1.00 | 0.97 |

In the case of predicting religiosity in different countries, handling the imbalanced data is a necessary step before classification. Resampling is an effective and computationally efficient approach to address this issue. Random forest is a powerful algorithm that can be used to classify individuals according to their level of religiosity. WVS contains a large number of variables that could potentially be relevant for determining an individual's religiosity, such as age, gender, education level, income, and country of origin, which makes it suitable for training with random forest. Additionally, random forest is robust to outliers and missing data, which is important when working with real-world datasets.

## 4 Conclusion

In this paper, we analyzed the WVS dataset to explore the key factors of religiosity in societies. By using the random forest, we can gain insights into the factors that are most strongly associated with religious beliefs and practices around the world and how these factors interact with one another. The results of variable importance analysis using random forest suggest that age and income are the two most important variables, which is consistent with the findings of research in sociology. We also investigated the correlation between religiosity with political views and life satisfaction in the context of different countries. And we interpreted the results based on fundamental theses in sociology, such as secularization theory, existential security theory, rational choice theory, and inequality and stratification. Further quantitative analysis of other variables such as education, marital status, and social class remains a future research problem to provide a more in-depth understanding of environmental factors on personal beliefs and values and their association with the community.

We also explored the problem of imbalanced learning in the context of the classification of religious beliefs using historical WVS data. We aimed to balance the data using multiple resampling methods, and the results from training and testing the random forest on the balanced dataset prove the effectiveness of this approach.

This study is an application of machine learning in sociology and human sciences. The data processing and classification methods used in this study are great tools for gaining insights into key factors of religiosity in societies using survey data. For future research, we recommend expanding the study to other cultural contexts, as our study is limited to 30 countries. Examining the causal factors behind religiosity is also beneficial in understanding why certain beliefs and attitudes are associated with higher levels of religiosity.

In conclusion, this study demonstrates the broad applications of machine learning models in social sciences. It exemplifies the use of data processing and classification methods on survey data by analyzing a sample of 30 countries participating in WVS with a variety of cultural and socioeconomic backgrounds. Although machine learning is not a panacea, it can be used in conjunction with other research methods to provide a comprehensive understanding of human beliefs, values, and religion. The results of our analysis highlight the importance of personal factors in determining the individual's level of religiosity. Such studies can provide valuable information for policymakers and organizations seeking to understand the role of religion in society and develop targeted interventions.